\documentclass[runningheads]{llncs}

\usepackage{eccv}

\usepackage{eccvabbrv}

\usepackage[dvipsnames]{xcolor}
\usepackage{booktabs}
\usepackage{multirow}
\usepackage{multicol}
\usepackage{color, colortbl}
\usepackage{graphics}
\usepackage{algpseudocode}
\usepackage{algorithm}
\usepackage{cellspace}
\usepackage{arydshln}
\usepackage{caption}
\usepackage{subcaption}
\definecolor{Gray}{gray}{0.9}

\usepackage[accsupp]{axessibility}  

\usepackage{hyperref}

\usepackage{orcidlink}
\newcommand*\samethanks[1][\value{footnote}]{\footnotemark[#1]}

\begin{document}

\title{Towards Model-Agnostic Dataset Condensation by Heterogeneous Models} 

\titlerunning{Heterogeneous Models Dataset Condensation}

\author{Jun-Yeong Moon\orcidlink{0000-0002-0543-2588}\and
Jung Uk Kim\thanks{Corresponding authors}\orcidlink{0000-0003-4533-4875}\and
Gyeong-Moon Park\samethanks\orcidlink{0000-0003-4011-9981}
}
\authorrunning{J.-Y. Moon et. al.}

\institute{Kyung Hee University, Yongin, Republic of Korea \\
\email{\{moonjunyyy, ju.kim, gmpark\}@khu.ac.kr}}

\maketitle

\begin{abstract}
The advancement of deep learning has coincided with the proliferation of both models and available data.
The surge in dataset sizes and the subsequent surge in computational requirements have led to the development of the Dataset Condensation (DC).
While prior studies have delved into generating synthetic images through methods like distribution alignment and training trajectory tracking for more efficient model training, a significant challenge arises when employing these condensed images practically.
Notably, these condensed images tend to be specific to particular models, constraining their versatility and practicality.
In response to this limitation, we introduce a novel method, Heterogeneous Model Dataset Condensation (HMDC), designed to produce universally applicable condensed images through cross-model interactions.
To address the issues of gradient magnitude difference and semantic distance in models when utilizing heterogeneous models, we propose the Gradient Balance Module (GBM) and Mutual Distillation (MD) with the Spatial-Semantic Decomposition method.
By balancing the contribution of each model and maintaining their semantic meaning closely, our approach overcomes the limitations associated with model-specific condensed images and enhances the broader utility.
The source code is available in \url{https://github.com/KHU-AGI/HMDC}.
\end{abstract}

\section{Introduction}
\label{sec:intro}

In recent years, deep learning \cite{rumelhart1985learning, lecun1998gradient, lecun2015deep} has demonstrated a remarkable surge in both effectiveness and applicability across diverse domains \cite{ vaswani2017attention, goodfellow2014generative, dosovitskiy2020image, ho2020denoising}.
With the increasing depth and complexity of models, the need for substantial datasets has become imperative to sustain their performance and forestall overfitting \cite{cubuk2018autoaugment, karystinos2000overfitting}.
Yet, the challenges extend beyond the mere acquisition of huge datasets to management and efficient utilization.
In this context, techniques such as dataset distillation (DD) \cite{wang2018dataset} or dataset condensation (DC) \cite{zhao2020dataset} have emerged, aiming to address these challenges by offering more efficient data management strategies.
These methods not only enable the selection of core-sets \cite{10.1145/1008731.1008736, 10.1145/1007352.1007400, feldman2020turning} capable of maintaining the original performance using a whole dataset but also facilitate a dramatic reduction in dataset size through the creation of synthetic data \cite{wang2018dataset, nguyen2021dataset, cazenavette2022dataset, zhao2020dataset, zhou2022dataset, kim2022dataset, wang2022cafe, zhao2023improved, cazenavette2023generalizing, Liu_2023_ICCV} that accurately represents the original dataset.

\begin{figure}[t]
    \centering
    \caption{Accuracy plots illustrating the performance of different models trained using images generated by recent dataset condensation methods on the CIFAR-10 dataset with an IPC10 setting. Each bar signifies a performance comparison relative to randomly selected images on 10 images per class, with the initial state of each method identical to that of the random image. Notably, the methods exhibit over-condensation on ConvNet, resulting in performance degradation on other models.}
    \includegraphics[width=1.0\textwidth]{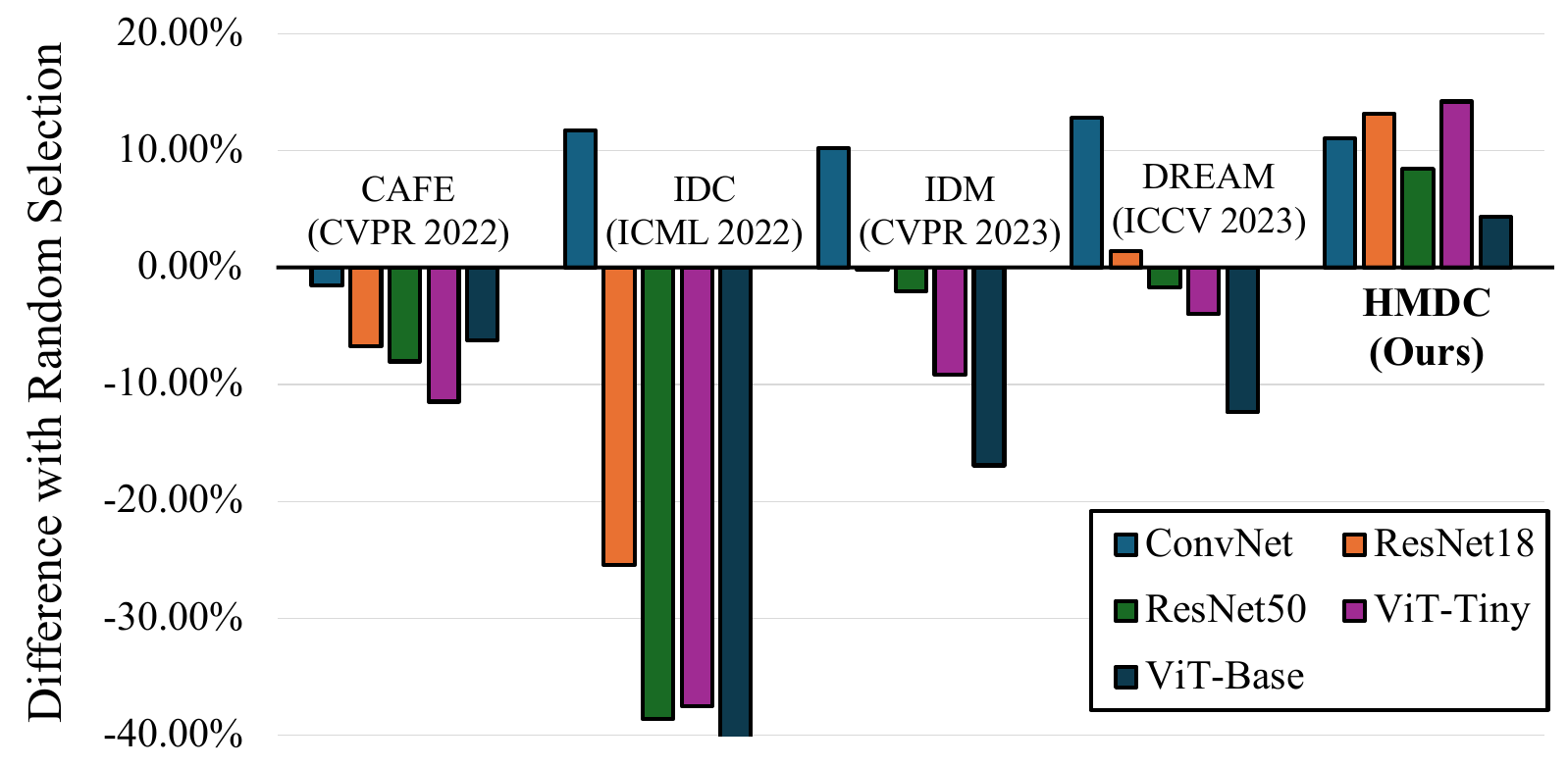}
    \label{fig:dependency}
\end{figure}

Traditionally, dataset condensation methods have employed a compact 3-layered model named ConvNet \cite{zhao2020dataset} for the condensation process \cite{cazenavette2022dataset, wang2022cafe, zhao2023improved, Liu_2023_ICCV}.
The standard evaluation method for assessing the performance of condensed images has been the \textit{ConvNet-to-ConvNet} test, where condensation and evaluation are conducted on ConvNet.
Some studies have delved into assessing the general performance of condensed images on other shallow models, such as 3-layered MLP or AlexNet \cite{krizhevsky2012imagenet}.
However, as illustrated in Figure \ref{fig:dependency}, the effectiveness of generated images diminishes when applied to widely used models like ResNet \cite{he2016deep} and Vision Transformer (ViT) \cite{dosovitskiy2020image}.
This indicates that synthetic images are over-condensed on ConvNet, showing a high model dependence.
This dependency significantly constrains the versatility of condensed images.
Introducing a completely new model necessitates training a new model and generating a new condensed image, implying that training data must be stored in some way.
Consequently, there is a need for a model-agnostic benchmark, not limited to \textit{ConvNet-to-ConvNet}, that operates independently of the specific model.

The primary challenge in achieving model-agnostic dataset condensation lies in identifying the common characteristics within a model and devising an effective method for their extraction.
It is difficult to distinguish between features with general recognition information and those excessively tailored to a particular model.
To address this challenge, for the first time, we introduce a novel approach that utilizes two models to extract generalized knowledge without bias towards any specific model. To this end, we propose a novel dataset condensation method, Heterogeneous Model Dataset Condensation (HMDC), leveraging heterogeneous models to extract common features that are more universally applicable.

When naively employing two heterogeneous models simultaneously for dataset condensation, two difficult issues arise.
Firstly, there is a problem of gradient magnitude difference, characterized by significant differences in the size of the gradient provided to the synthetic image due to structural or depth variations between the two models.
This discrepancy can lead to the disregard of one model or the failure of the image to converge.
To alleviate this, we present a Gradient Balance Module (GBM), which accumulates the gradient magnitude of each optimization target, to control the magnitude of the loss.
Thus, even if two models have different structures, they can have a similar impact on the synthetic image.

Another challenge arises in the semantic distance resulting from different knowledge between models.
As the two models undergo learning, they converge towards optimal points specific to other models, leading to a growing of semantic distance and instability in image convergence.
To tackle this, we propose a Mutual Distillation (MD) by Spatial-Semantic Decomposition of the two models and feature matching throughout the process.
This process enables consistent updates of synthetic images regardless of the model, avoiding over-condensation on any particular model by obtaining information from different models.
This characteristic makes our method effective in a model-agnostic setting.

Our main contributions can be summarized as follows:
\begin{itemize}
  \item For the first time, we present Heterogeneous Model Dataset Condensation (HMDC) for model-agnostic dataset condensation, which resolves the over-condensation issue to a specific model.
  \item To facilitate the convergence of synthetic images, we propose the Gradient Balance Module to control a gap between the gradient magnitudes of heterogeneous models.
  \item We propose Mutual Distillation by Spatial-Semantic Decomposition feature matching of heterogeneous models to fill in the semantic distance between models.
  \item From the extensive experiments, we demonstrate that our condensed images consistently show great performances from shallow models to widely used large models.
\end{itemize}
\section{Related Work}
\label{sec:rw}

\vspace{-1mm}
\subsection{Dataset Condensation}
\vspace{-1mm}
\label{sec:rw:dc}

Through the exploration of various core-set selection methodologies \cite{10.1145/1008731.1008736, 10.1145/1007352.1007400, feldman2020turning}, it has become evident that synthesized images generated through optimization procedures exhibit greater efficacy compared to direct utilization of real images.
Optimization techniques for generating synthetic images primarily fall into two categories: those concerned with tracking training trajectories and those focused on aligning feature distributions. 
The trajectory tracking approach involves aligning the gradients \cite{zhao2020dataset, zhou2022dataset, kim2022dataset, Liu_2023_ICCV} between real and synthetic images during the training process, or adapting the weights of the model trained on synthetic images to resemble the model trained on real images \cite{cazenavette2022dataset}.
These techniques aim to generate synthetic images based on their influence on the model's learning process.
Another stream of condensation methods emphasizes feature distribution matching at intermediate layers \cite{wang2022cafe}, or output distribution alignment with the synthetic images \cite{wang2018dataset, nguyen2021dataset, zhou2022dataset, zhao2023improved}.
These approaches aim to generate synthetic images by emphasizing feature similarity within the synthetic image but they still depend on utilizing an intermediate training state of the model.
In this work, we use the gradient matching method, which has shown better performance in previous studies \cite{zhao2020dataset, zhou2022dataset, kim2022dataset, Liu_2023_ICCV}. 
In the existing studies \cite{cazenavette2022dataset, wang2022cafe,zhao2023improved, Liu_2023_ICCV}, synthetic images were typically generated using small 3-layer models, resulting in images that exhibited limited compatibility with other models.
In contrast to conventional approaches, our method employs two distinct models to generate a balanced condensed image that avoids being overly biased toward either model, addressing limitations observed in earlier approaches.

\vspace{-1mm}
\subsection{Knowledge Distillation}
\vspace{-1mm}

Knowledge Distillation (KD), as introduced by Hinton et al. \cite{hinton2015distilling}, is a technique in machine learning where a smaller model known as the student model trains to replicate the behavior of a larger model which is the teacher model. This process is done to transfer the knowledge and generalization capabilities of the teacher model to the smaller and more efficient student model. This transfer of knowledge in KD is achieved by aligning what is often referred to as dark knowledge, which can manifest as either logits \cite{furlanello2018born, yang2019snapshot, mirzadeh2020improved, zhao2022decoupled} or features \cite{heo2019comprehensive, heo2019knowledge, huang2017like, kim2018paraphrasing}.

While the primary objective of KD is to create a more efficient smaller model that performs similarly to a larger one, in this study, we leverage KD to specifically reduce the semantic distance between models. This approach enables learning from a single image through the knowledge of two distinct models, thereby ensuring stable learning without the risk of collapse.

\vspace{-1mm}
\subsection{Utilization of Heterogeneity}
\vspace{-1mm}

Previous methods for adjusting loss or gradient have primarily focused on using the uncertainty \cite{kendall2018multi} or norm of the gradient \cite{chen2018gradnorm} to balance multi-task learning or employing multiple adaptors for domain-robust models \cite{li2021universal}. In this work, we claim that while there is a single task, heterogeneity is necessary to solve it effectively. We decompose the features of the image model into spatial and semantic information, allowing us to simultaneously leverage the knowledge of two models with distinct features. By accumulating the gradient norm, we can identify differences in the average gradient and appropriately scale it to inject more general features into the synthetic images, thereby compensating for the imbalance in learning caused by the structures of models.

\section{Method}
\label{sec:method}

\begin{figure}[t!]
    \centering
    \caption{Diagram of Heterogeneous Model Dataset Condensation (HMDC), where two distinct models are employed for feature extraction. These features undergo dimension adjustment through Spatial-Semantic Decomposition, a critical step facilitating Mutual Distillation, and enhancing knowledge sharing between the two models. Throughout the dataset condensation process, the compensatory Gradient Balance Module comes into play, mitigating gradient variations inherent to different models. This module ensures the extraction of general knowledge by harmonizing gradient magnitudes, thus contributing to a more universally applicable condensation process.}
    \includegraphics[width=\textwidth]{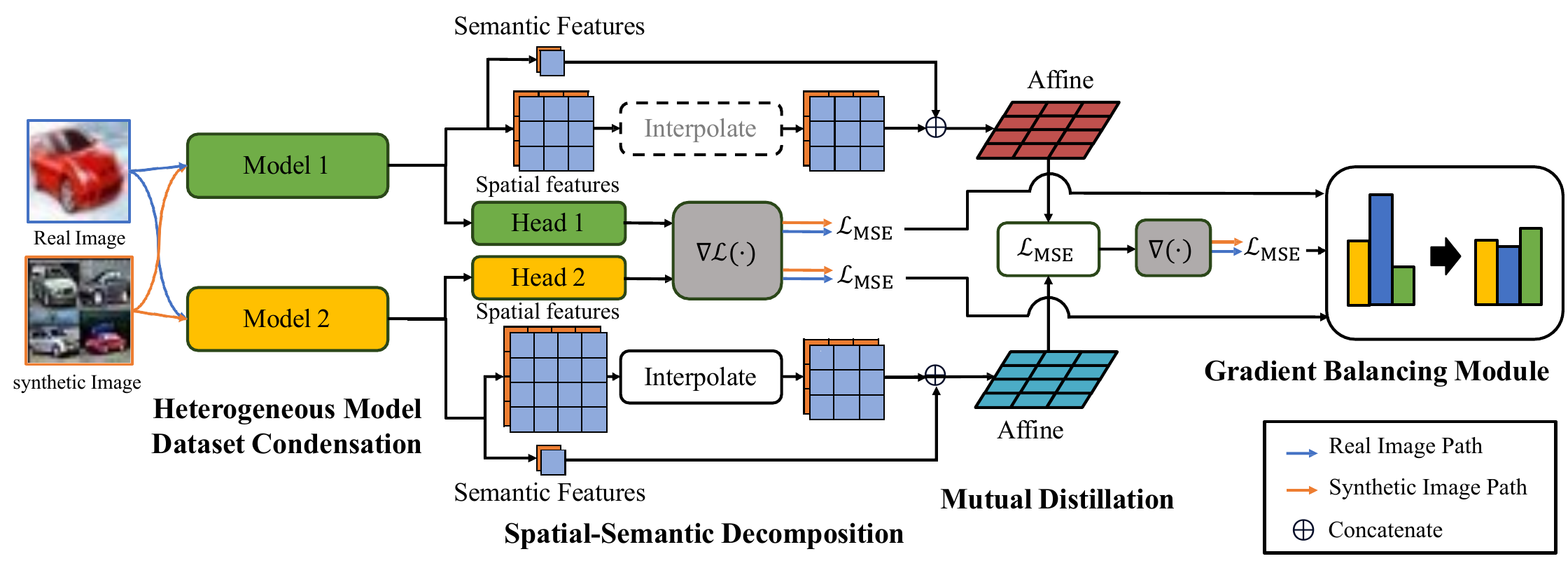}
    \label{fig:mainfig}
\end{figure}

\subsection{Problem Formulation}
\label{sec:method:form}
Dataset Condensation (DC) \cite{zhao2020dataset} is an approach that aims at creating a synthetic dataset denoted as $\mathcal{S} = {(\textbf{x}_i, y_i)}^{|\mathcal{S}|}_{i=1}$ from the complete training data $\mathcal{T} = {(\textbf{x}_i, y_i)}^{|\mathcal{T}|}_{i=1}$, where $|\mathcal{S}| \ll |\mathcal{T}|$. This synthetic dataset, $\mathcal{S}$, is designed to train a model to a performance level comparable to what could be achieved with the original data. The dataset $\mathcal{S}$ can encompass a subset of $\mathcal{T}$. However, recent research has unveiled that the utilization of synthetic data yields superior performance \cite{wang2018dataset, nguyen2021dataset, cazenavette2022dataset, zhao2020dataset, zhou2022dataset, kim2022dataset, wang2022cafe, zhao2023improved, cazenavette2023generalizing, Liu_2023_ICCV}. In many cases, the central challenge in dataset condensation revolves around the determination of the optimization target, denoted as $\phi(\textbf{x},y)$, for the condensed image set $\mathcal{S}$. This can be formally expressed as an optimization problem:
\begin{align}
\mathcal{S} = \arg \underset{\mathcal{S}}{\mathrm{min}} \sum_{i=1}^{|\mathcal{B}|} \, \mathcal{D}(\phi(\textbf{x}^t_i, y^t_i), \phi(\textbf{x}^s_i, y^s_i)), ~ {(\textbf{x}^t_i, y^t_i)}^{|\mathcal{B}|}_{i=1} \sim \mathcal{T},\;{(\textbf{x}^s_i, y^s_i)}^{|\mathcal{B}|}_{i=1} \sim \mathcal{S}, \label{eq:condense}
\end{align}
where $\mathcal{D}(\cdot,\cdot)$ represents a matching function, such as Euclidean or cosine distance; commonly, Mean Squared Error (MSE) is used, and $\mathcal{B}$ represents a mini-batch. $\phi$ is typically associated with features, gradients, or weights of the model.
For model-agnostic DC, we utilize a gradient-based method that involves two heterogeneous models, enabling the condensed image to acquire more generalized knowledge. 
In the following sections, we introduce Heterogeneous Model Dataset Condensation (HMDC) along with methods to find an appropriate optimization target.

\subsection{Heterogeneous Model Dataset Condensation}
\label{sec:method:Hete}
Traditional methods for gradient-based dataset condensation typically utilize the gradient of the model's cross-entropy loss as the optimization target \cite{zhao2020dataset, zhao2023improved}, denoted as $\phi(\textbf{x},y)$. This is expressed mathematically as:
\begin{align}
\phi(\textbf{x},y) = \nabla \mathcal{L}_{\mathrm{CE}}(f_\theta(\textbf{x}),y).
\label{eq:gradient}
\end{align}
Here, $f_\theta$ represents a model parameterized by $\theta$ and $\mathcal{L}_\mathrm{CE}$ is the cross-entropy loss. However, these approaches exhibit a model-dependent nature since they focus on training the model's path on the image. In contrast, our HMDC seeks to extract the common features by concurrently considering the training paths of two models, $f_{\theta_1}$ and $f_{\theta_2}$, where two models complement their features each other as illustrated in Figure \ref{fig:mainfig}. This is expressed mathematically  as:
\begin{align}
\mathcal{S} = \arg \underset{\mathcal{S}}{\mathrm{min}} &\left\{\sum_{i=1}^{|\mathcal{B}|} \, \mathcal{D}\left(\nabla \mathcal{L}_{\mathrm{CE}}\left(f_{\theta_{1}}\left(\textbf{x}^t_i\right), y^t_i\right), \nabla \mathcal{L}_{\mathrm{CE}}\left(f_{\theta_{1}}\left(\textbf{x}^s_i\right), y^s_i\right)\right) \right.\nonumber\\
&\left.+ \sum_{i=1}^{|\mathcal{B}|} \, \mathcal{D}\left(\nabla \mathcal{L}_{\mathrm{CE}}\left(f_{\theta_{2}}\left(\textbf{x}^t_i\right), y^t_i\right), \nabla \mathcal{L}_{\mathrm{CE}}\left(f_{\theta_{2}}\left(\textbf{x}^s_i\right),y^s_i\right)\right)\right\}\nonumber\\
=\arg \underset{\mathcal{S}}{\mathrm{min}}&\left\{\sum_{i=1}^{|\mathcal{B}|}\mathcal{L}^1+\sum_{i=1}^{|\mathcal{B}|}\mathcal{L}^2\right\}, 
\label{eq:dual}
\end{align}
where $\mathcal{L}^1$ and $\mathcal{L}^2$ mean the optimization target about $f_{\theta_1}$ and $f_{\theta_2}$ respectively, in this case, the mean squared error between gradients generated by synthetic and real images in each model. To enhance performance, additional regularization terms can be introduced as $\mathcal{L}^3, ..., \mathcal{L}^k$. This formulation envisions a dataset condensation method that straightforwardly utilizes two models. However, two key challenges arise the gradient magnitude difference and semantic distance between the models. We address these challenges with our novel Gradient Balance Module (GBM) and Mutual Distillation (MD) with Spatial-Semantic Decomposition (SSD), which are described next.

\subsection{Gradient Balance Module}
\label{sec:method:gbm}
The optimization of Eq. \ref{eq:dual} faces new challenges due to a significant disparity in the gradient magnitude among $\nabla\mathcal{L}^1, \nabla\mathcal{L}^2, ... , \nabla\mathcal{L}^k$.
This discrepancy can stem from various factors, such as differences in model structure, depth, and feature dimensionality.
Furthermore, if there are additional optimization targets, a hyperparameter search becomes necessary.
The varying magnitudes in these gradients could potentially lead to the neglect of one side or hinder convergence.

To address this, we propose the gradient balance module as illustrated in Figure \ref{fig:mainfig}, which sets up an accumulator to store the size of the gradient from each optimization target. The accumulator $\mathcal{A} = \left\{a_1, a_2, ..., a_k\right\} \in \mathbb{R}^k$  is expressed as:
\begin{align}    \mathcal{A}=\left[\sum_{s=1}^S\max\left(\left|\nabla\mathcal{L}^1(s)\right|\right), \sum_{s=1}^S\max\left(\left|\nabla\mathcal{L}^2(s)\right|\right), ... , \sum_{s=1}^S\max\left(\left|\nabla\mathcal{L}^k(s)\right|\right)\right],
\label{eq:accum}
\end{align}
where $S$ represents the total number of optimization steps, an aforementioned $k$ is the number of the optimization targets, $\mathrm{max}\left(\cdot\right)$ represents the maximum scalar value in a given tensor, and there are $k$ optimization targets. During the optimization, the reciprocal of the normalized accumulator is multiplied to ensure a similar gradient amplitude for each element:
\begin{align}
    \mathcal{S}=\arg \underset{\mathcal{S}}{\mathrm{min}} \left\{\left[\mathcal{L}^1, \mathcal{L}^2, ...,\mathcal{L}^k\right]\cdot \min(\mathcal{A})\mathcal{A}^\mathrm{R}\right\},
\label{eq:normalize}
\end{align}
where $\mathrm{min}\left(\cdot\right)$ represents the minimum scalar value in a given tensor, $\mathcal{A}^\mathrm{R}$ represents element-wise reciprocal of vector $\mathcal{A}$.
In our study, we utilize three distinct losses as optimization targets.
To address the imbalance between images in the update and prevent image collapse, we normalize the gradient of each synthetic image.
To manage the computational intensity associated with additional gradient computations, we apply an accumulation strategy once in a while, \textit{e.g.}, by sampling once every 10 steps.

\subsection{Mutual Distillation with Spatial-Semantic Decomposition}
\label{sec:method:md}
Eq. \ref{eq:dual} designed to track the learning path of each model through the gradient matching.
However, if the meanings of the two features differ significantly, it impedes the effective learning of synthetic images and hinders their convergence.
To mitigate this, we propose a new mutual distillation loss that restricts the semantic divergence between the two models in training processes.
On top of that, since two different models vary in depth, dimensionality, and the number of features, effective distillation between them requires careful consideration. 

To address this challenge, we introduce a Spatial-Semantic Decomposition (SSD) to preserve the semantics of the features to the maximum extent possible.
It decomposes the features of a model into spatial and semantic parts and transforms them accordingly, allowing you to compare the semantics of two models with a linear projection.
We designate the features related to classification as semantic features, representing the entire image, and those characterizing each spatial location as spatial features.
As spatial features can be represented in an image-like form, features of varying sizes can be aligned using bilinear interpolation, simplifying the alignment process without requiring complex transformations.
For easy understanding, here we explain our proposed method using two example models below Vision Transformer (ViT) \cite{dosovitskiy2020image} and Convolutional Neural Network (CNN) \cite{lecun1998gradient}. Note that two heterogeneous models can vary, not limited to them.\\

\noindent\textbf{ViT Architecture.} First, in the ViT architecture, we utilize a class token (CLS) attached to the front of input tokens as semantic features, and the other image tokens as spatial features. Given the feature from every layer of ViT in the dimension of  $\mathcal{F}_{\mathrm{ViT}}\in \mathbb{R}^{n \times (w_\mathrm{ViT}h_\mathrm{ViT}+1) \times d_\mathrm{ViT}}$:
\begin{align}
    \mathcal{F}_{\mathrm{ViT}}^{\mathrm{sementic}} &= \mathcal{F}_{\mathrm{ViT}} \left[\,:, 0, :\, \right] \;\;\in \mathbb{R}^{n \times 1\times d_\mathrm{ViT}},\nonumber\\
    \mathcal{F}_{\mathrm{ViT}}^{\mathrm{spatial}} &= \mathcal{F}_{\mathrm{ViT}} \left[\,:, 1:,:\, \right] \in \mathbb{R}^{n \times (w_\mathrm{ViT}\times h_\mathrm{ViT}) \times d_\mathrm{ViT}},
\end{align}
where $n$ means the number of layers, $w_\mathrm{ViT}$ and $h_\mathrm{ViT}$ means the number of tokens generated by the ViT when it performs patch embedding to generate image tokens in the horizontal and vertical directions, respectively. The CLS token is attached to the front, and $d_\mathrm{ViT}$ is the dimension of each token. Given the variations in the size of spatial features across layers and models, standardization is achieved through bilinear interpolation. To address differences in feature dimensionality, we apply a learnable affine transformation. The dimension-aligned feature is expressed as:
\begin{align}
    \mathcal{F}_{\mathrm{1}} = &\left[W_1^1\left[\mathcal{F}_{\mathrm{ViT}}^{\mathrm{sementic}}[1];\underset{w \times h}{\mathrm{I}}\left(\mathcal{F}_{\mathrm{ViT}}^{\mathrm{spatial}}[1]\right)\right]+\textbf{b}^1_1;\nonumber\right.\\
    &\left....;W_n^1\left[\mathcal{F}_{\mathrm{ViT}}^{\mathrm{sementic}}[n];\underset{w \times h}{\mathrm{I}}\left(\mathcal{F}_{\mathrm{ViT}}^{\mathrm{spatial}}[n]\right)\right]+\textbf{b}^1_n\right],
\label{eq:vit}
\end{align}
where $n$ is the number of layers in ViT, $W^1_1,..., W^1_n\in\mathbb{R}^{d\times{d}_\mathrm{ViT}}$, $\textbf{b}^1_1,..., \textbf{b}^1_n\in\mathbb{R}^{d}$, and $\underset{w\times h}{\mathrm{I}}$ refers to bilinear interpolation into the size of $w \times h$.
As a result, we get the feature of $\mathbb{R}^{n \times (wh+1) \times d}$.
$w$, $h$, and $d$ mean a target interpolation width, height, and dimension respectively. We use dimension and interpolation size into smaller values between two models, which is found to be empirically better.
For computational convenience, it is transposed to $\mathbb{R}^{(wh+1) \times d \times n}$.
\\

\noindent\textbf{CNN Architecture.} In the case of CNN features, inherently semantic features can be generated through mean pooling of spatial features. This process is mathematically expressed as:
\begin{align}
    \mathcal{F}_{\mathrm{CNN}} &= \{\mathcal{F}_{\mathrm{CNN}}^1,... ,\mathcal{F}_{\mathrm{CNN}}^m\},\quad
    \mathcal{F}_{\mathrm{CNN}}^l\in \mathbb{R}^{ d_l \times w_l \times h_l }.\\
    \mathcal{F}_{\mathrm{CNN}}^{\mathrm{sementic}} &= \left\{\frac{1}{w_1 h_1}\sum^{w_1}_{w=1}\sum^{h_1}_{h=1}\mathcal{F}_{\mathrm{CNN}}^1\left[\,:,w,h\,\right],...,\frac{1}{w_m h_m}\sum^{w_m}_{w=1}\sum^{h_m}_{h=1}\mathcal{F}_{\mathrm{CNN}}^m\left[\,:,w,h\,\right]\right\},\nonumber\\
    \mathcal{F}_{\mathrm{CNN}}^{\mathrm{spatial}} &= \mathcal{F}_{\mathrm{CNN}},\qquad\qquad\qquad\qquad\qquad\quad\,
\end{align}
where m is the number of layers in CNN. As observed in the case of ViT, we standardize dimension and feature size via bilinear interpolation and affine transformation. This process is expressed as:
\begin{align}
    \mathcal{F}_{\mathrm{2}} = \left[W_1^2\left[\mathcal{F}_{\mathrm{CNN}}^{\mathrm{sementic}}[1];\underset{w \times h}{\mathrm{I}}\left(\mathcal{F}_{\mathrm{CNN}}^{\mathrm{spatial}}[1]\right)\right]+\textbf{b}^2_1;\nonumber\right.\\...;
    \left.W_m^2\left[\mathcal{F}_{\mathrm{CNN}}^{\mathrm{sementic}}[m];\underset{w \times h}{\mathrm{I}}\left(\mathcal{F}_{\mathrm{CNN}}^{\mathrm{spatial}}[m]\right)\right]+\textbf{b}^2_m\right],
\label{eq:cnn}
\end{align}
where $W^2_1\in\mathbb{R}^{d\times{d_1}},..., W^2_m\in\mathbb{R}^{d\times{d_{m}}}$, and $\textbf{b}^2_1,..., \textbf{b}^2_m\in\mathbb{R}^{d}$. Finally we get the feature of $\mathbb{R}^{m \times (wh+1) \times d}$. And, it is transposed to $\mathbb{R}^{(wh+1) \times d \times m}$.

The aligned dimensionality and the number of features are currently the same, but the number of layers in the two models is still different, \textit{i.e.},
$\mathcal{F}_{\mathrm{1}}\in\mathbb{R}^{(wh+1) \times d\times n}$, $\mathcal{F}_{\mathrm{2}}\in\mathbb{R}^{(wh+1) \times d \times m}$.  We further match the number of layers by using an ${n\times m}$ matrix $M_\mathrm{layer}\in\mathbb{R}^{n\times m}$, where $n$ represents the number of layers in one model and $m$ denotes the number of layers in the other. Softmax is applied to this matrix for layer selection. We call this matching process as Spatial-Semantic Decomposition (SSD), where the matched features can be expressed as follows:
\begin{align}
\left\{ 
  \begin{array}{ c l }
    \mathcal{F}_1, \mathcal{F}_2\,\cdot\mathrm{softmax}(M^\mathrm{T}_\mathrm{layer})   & \quad m > n \\
    \mathcal{F}_1\,\cdot\mathrm{softmax}(M_\mathrm{layer}),\mathcal{F}_2  & \quad \textrm{otherwise}
  \end{array}
\right.\;.
\label{eq:layer}
\end{align}
Here, we align the number of layers to the smaller one, which is found to be empirically better.
This gives us two $\mathbb{R}^{(wh+1) \times d\times n}$ of dimension-aligned features when $m>n$ or $\mathbb{R}^{(wh+1) \times d \times m}$ dimension of feature otherwise.
Both the feature affine matrixes ($W^1$ and $W^2$) and the layer matching matrix ($M_\mathrm{layer}$) undergo training to ensure feature alignment between the models at each step.
Note again that we give an example using CNN and ViT, but our method can be extended to any model with spatial features.

The Spatial-Semantic Decomposition method enables a comparison of the distinct features of two models. 
Throughout the training process, we propose Mutual Distillation (MD) to align the meanings of these features, aiming to make them similar knowledge across both models.
The training loss for each model $f_1$ and $f_2$ is expressed as:
\begin{align}
    &\mathcal{L}_\mathrm{MD}\left(\textbf{x}\right) = \mathrm{MSE}(\mathrm{SSD}\left(f_{\theta_1}\left(\textbf{x}\right),f_{\theta_2}\left(\textbf{x}\right)\right)),\nonumber\\
    &\mathcal{L}_{f_{1}}=\mathcal{L}_{\mathrm{CE}}\left(f_{\theta_1}\left(\textbf{x}\right),y\right) + \mathcal{L}_\mathrm{MD}(\textbf{x}),\;\,\nonumber\\
    &\mathcal{L}_{f_{2}}=\mathcal{L}_{\mathrm{CE}}\left(f_{\theta_2}\left(\textbf{x}\right),y\right) + \mathcal{L}_\mathrm{MD}(\textbf{x}).\;\,
\label{eq:distill}
\end{align}
 To guide the image in learning intermediate information, the Mutual Distillation loss serves as an additional regularization term for the synthetic image. Through this process, each semantic aspect of the model and the learning path of the synthetic image are guided, achieving a balance between the two models and enhancing generality in the results. Consequently, the number of optimization target $k$ for condensed images is $3$ in this case. Total loss function for condensed images can be defined as a vector inner product as follows:
\begin{align}
    \textbf{L} &= \left[\mathcal{L}^1,\mathcal{L}^2,\mathrm{MSE}\left(\nabla\mathcal{L}_{\mathrm{MD}}(\textbf{x}^t),\nabla\mathcal{L}_{\mathrm{MD}}(\textbf{x}^s)\right)\right],\nonumber\\
    \mathcal{L}_{\mathrm{target}}&=\textbf{L} \cdot \min\left(\mathcal{A}\right) \mathcal{A}^\mathrm{R} \nonumber\\
    &=\frac{\min\left(\mathcal{A}\right)}{a_1}\mathcal{L}_1 + \frac{\min\left(\mathcal{A}\right)}{a_2}\mathcal{L}_2  + \frac{\min\left(\mathcal{A}\right)}{a_3}\mathrm{MSE}\left(\nabla\mathcal{L}_{\mathrm{MD}}(\textbf{x}^t),\nabla\mathcal{L}_{\mathrm{MD}}(\textbf{x}^s)\right),
\label{eq:target}
\end{align}
where $a_1, a_2,$ and $a_3$ is current value in the accumulator $\mathcal{A}$ part of the Gradient Balance Module. This allows us to learn by considering the semantic distance between two models when training a model and when training an image, and to extract more general knowledge from it.

\section{Experiments}

\subsection{Implementation Details}
\label{sec:exp:detail}

In this study, we performed a comprehensive comparative analysis, assessing the effectiveness of our proposed method against cutting-edge gradient matching techniques, IDC \cite{kim2022dataset} and DREAM \cite{Liu_2023_ICCV}. To broaden our evaluation scope to distribution matching, we selected CAFE \cite{wang2022cafe} and IDM \cite{zhao2023improved} as benchmark methodologies.

To ensure a fair and standardized comparison, we employed a consistent augmentation strategy across all methods. This strategy encompassed a sequence of color modifications, cropping, and either Cutout \cite{devries2017improved} or CutMix \cite{yun2019cutmix}, aligning with the recommended practices outlined in IDC \cite{zhao2023improved}.
Also, we use multi-formation which is used in IDC and IDM \cite{zhao2023improved} for every method.
Nonetheless, it appears that while this robust augmentation technique enhances the performance of the gradient-based method, it adversely affects the distribution matching method, resulting in a decline in performance. In the images generated during the process, a noticeable inclination to produce corrupted images under intense augmentation was observed.

To measure how the condensed image effectively trains the large model, We conducted experiments on CIFAR10 \cite{krizhevsky2009learning} on Images Per Class (IPC) 1, 10, 50. 
Our assessment followed the experimental procedures detailed in each referenced paper.
For evaluation, we utilized ConvNet, ViT-tiny \cite{dosovitskiy2020image}, ResNet18 \cite{he2016deep}, ViT-small, ResNet50, ResNet101, and ViT-base models, each with specific learning rates ($0.01$, $0.001$, $0.001$, $0.001$, $0.001$, $0.0001$, and $0.0001$, respectively).
The best scores were measured as performance metrics during the training of the models for a consistent duration of 2,000 epochs. Because training a large model on a small dataset makes it difficult to compare performance, every model in Table \ref{table:main_table:CIFAR10} except ConvNet was pre-trained on the ImagiNet-1K \cite{deng2009imagenet} dataset.

\begin{table}[t!]
\caption{Experimental results of dataset condensation methods on CIFAR-10. HMDC is the best or second-best performer in most cases.}
\label{table:main_table:CIFAR10}
\scalebox{0.57}{
\begin{tabular}{Sc|Sc|ScScScScSc:ScScScSc:Sc}
\hline
\multirow{3}{*}[-2pt]{IPC} & \multirow{3}{*}[-2pt]{Methods} & \multicolumn{10}{Sc}{Models (\#Params)}\\ \cline{3-12}
&&
\multirow{2}{*}[-.75pt]{\begin{tabular}[c]{@{}c@{}}ConvNet\\(0.3M)\end{tabular}} &
\multirow{2}{*}[-.75pt]{\begin{tabular}[c]{@{}c@{}}ResNet18\\(11M)\end{tabular}} &
\multirow{2}{*}[-.75pt]{\begin{tabular}[c]{@{}c@{}}ResNet50\\(22M)\end{tabular}} &
\multirow{2}{*}[-.75pt]{\begin{tabular}[c]{@{}c@{}}ResNet101\\(43M)\end{tabular}}&
\multirow{2}{*}[-.75pt]{\begin{tabular}[c]{@{}c@{}}CNN\\Average\end{tabular}} &
\multirow{2}{*}[-.75pt]{\begin{tabular}[c]{@{}c@{}}ViT-tiny\\(5.5M)\end{tabular}} &
\multirow{2}{*}[-.75pt]{\begin{tabular}[c]{@{}c@{}}ViT-small\\(21M)\end{tabular}} &
\multirow{2}{*}[-.75pt]{\begin{tabular}[c]{@{}c@{}}ViT-base\\(86M)\end{tabular}} &
\multirow{2}{*}[-.75pt]{\begin{tabular}[c]{@{}c@{}}ViT\\Average\end{tabular}} &
\multirow{2}{*}[-.75pt]{\begin{tabular}[c]{@{}c@{}}Average\end{tabular}} \\
&&&&&&&&& \\ \hline

\multirow{6}{*}[-6pt]{1}
 & \multicolumn{1}{Sc|}{Random}
 & 22.51±0.30 & \underline{39.39±3.43} & \underline{40.43±8.01} & \underline{51.17±1.04} & \underline{38.38±3.19} & \underline{32.49±3.59} & \textbf{58.33±1.42} & \underline{41.05±5.99} & \underline{43.96±3.66} & \underline{40.77±3.39} \\
 & \multicolumn{1}{Sc|}{CAFE}
 & 27.16±0.74 & 22.65±4.08 & 24.83±1.87 & 24.09±1.54 & 24.68±2.06 & 16.53±2.71 & 19.35±3.71 & 22.11±7.67 & 19.33±4.70 & 22.39±3.19 \\
 & \multicolumn{1}{Sc|}{IDM}
 & 37.51±0.32 & 17.50±2.36 & 16.07±1.20 & 11.47±0.57 & 20.64±1.11 & 12.85±0.74 & 11.40±1.57 & 12.12±4.27 & 12.12±2.19 & 16.99±1.57 \\
 & \multicolumn{1}{Sc|}{IDC}
 & 32.86±0.24 & 18.71±1.51 & 20.53±2.80 & 18.69±2.05 & 22.70±1.65 & 18.53±0.87 & 15.90±2.79 & 12.46±4.05 & 15.63±2.57 & 19.67±2.04 \\
 & \multicolumn{1}{Sc|}{DREAM}
 & \underline{38.76±0.47} & 29.75±2.40 & 28.99±2.45 & 33.24±8.03 & 32.68±3.33 & 19.86±1.00 & 21.82±1.96 & 22.00±4.50 & 21.23±2.49 & 27.77±2.97 \\ 
\rowcolor{lightgray}\cellcolor{white} & HMDC
 & \textbf{38.74±0.37} & \textbf{52.76±2.45} & \textbf{57.40±4.40} & \textbf{51.70±0.81} & \textbf{50.15±2.01} & \textbf{39.27±6.08} & \underline{56.21±8.00} & \textbf{49.46±10.9} & \textbf{48.31±8.31} & \textbf{49.36±4.71} \\ \hline
 
\multirow{6}{*}[-6pt]{10}
 & \multicolumn{1}{Sc|}{Random}
 & 36.45±0.12 & 56.59±4.86 & \underline{69.55±9.69} & \underline{82.03±2.06} & \underline{61.15±4.18} & \underline{59.41±9.20} & \textbf{90.11±1.25} & \underline{81.26±5.66} & \underline{76.93±5.37} & \underline{67.91±4.69} \\
 & \multicolumn{1}{Sc|}{CAFE}
 & 34.89±0.60 & 49.85±4.85 & 61.52±10.8 & 65.73±6.24 & 53.00±5.63 & 47.94±5.20 & 88.70±2.86 & 75.07±11.91 & 70.57±6.66 & 60.53±6.07 \\
 & \multicolumn{1}{Sc|}{IDM}
 & \underline{48.22±0.34} & 31.12±2.02 & 30.91±4.06 & 25.78±2.46 & 34.01±2.22 & 21.87±2.01 & 28.41±2.88 & 26.74±4.90 & 25.67±3.27 & 30.44±2.67 \\
 & \multicolumn{1}{Sc|}{IDC}
 & 46.67±0.15 & 56.35±1.92 & 67.53±2.77 & 60.94±4.48 & 57.87±2.33 & 50.23±8.38 & 68.67±8.50 & 64.32±12.32 & 61.07±9.73 & 59.24±5.50 \\
 & \multicolumn{1}{Sc|}{DREAM}
 & \textbf{49.23±0.16} & \underline{57.97±2.77} & 67.85±2.54 & 64.12±3.95 & 59.79±2.36 & 55.40±6.06 & 71.77±8.74 & 68.88±8.16 & 65.35±7.65 & 62.18±4.63 \\
 \rowcolor{lightgray}\cellcolor{white} & HMDC & 47.54±0.73 & \textbf{69.75±0.34} & \textbf{77.99±1.53} & \textbf{82.25±0.93} & \textbf{69.38±0.88} & \textbf{73.60±4.20} & \underline{89.02±1.42} & \textbf{85.58±1.77} & \textbf{82.73±2.46} & \textbf{75.10±1.56} \\ \hline
 
\multirow{6}{*}[-6pt]{50}
 & \multicolumn{1}{Sc|}{Random}
 & 45.47±0.39& \underline{73.55±1.02} &80.96±6.48&91.33±0.57&72.83±2.11&69.84±6.60&\textbf{96.14±0.06}&\textbf{93.98±1.99}&\textbf{86.65±2.88}&78.75±2.44 \\
 & \multicolumn{1}{Sc|}{CAFE}
 & 44.32±0.13&75.06±1.33& \underline{83.45±3.46} &\textbf{91.92±0.35}&\underline{73.69±1.32}&67.84±3.05&\underline{95.81±0.87}&\underline{92.86±2.92}&85.50±2.28&78.75±1.73 \\
 & \multicolumn{1}{Sc|}{IDM}
 & \underline{52.82±0.26} &64.01±0.47&72.60±0.49&67.15±12.6&64.14±3.45&53.55±4.13&77.52±15.4&74.56±11.6&68.54±10.4&66.03±6.42 \\
 & \multicolumn{1}{Sc|}{IDC}
 & 49.75±0.22&71.16±1.26&82.82±3.23&\underline{91.45±0.50}&73.80±1.30&\underline{70.37±1.52}&95.44±0.29&92.65±1.56&\underline{86.16±1.12}&\underline{79.09±1.22} \\
 & \multicolumn{1}{Sc|}{DREAM}
 & \textbf{52.90±0.16}&72.47±0.91&79.56±0.73&86.74±2.31&72.92±1.02&67.27±4.36&91.41±0.78&90.03±3.71&82.90±2.95&77.20±1.85 \\
\rowcolor{lightgray}\cellcolor{white} & HMDC 
 & 52.40±0.04 & \textbf{75.78±0.60} & \textbf{84.06±0.84} &89.08±0.30&\textbf{75.33±0.44}&\textbf{76.04±5.24}&90.91±0.54&91.45±0.68&86.13±2.15&\textbf{79.96±1.18} \\ \hline
\end{tabular}
}
\end{table}

In our approach, we adopt ConvNet and ViT-tiny as heterogeneous models. We configure the iteration parameter to $100$ and set the loop to iterate $100$ times for each iteration. Within each iteration, we perform updates to the image, update the model, and adjust the affine layer along with the layer-matching matrix. The learning rate of each model is $0.001$, and the affine layer and layer-matching matrix are $0.01$. Both use SGD optimizer \cite{robbins1951stochastic} as follows prior works. We set the batch size to 128.

\subsection{Results}
\label{sec:exp:result}
Table \ref{table:main_table:CIFAR10} presents a comprehensive performance comparison of the condensed images generated by each technique on CIFAR10. The proposed HMDC demonstrates commendable performance across most models, except ConvNet. 
Notably, other techniques generally exhibit inferior performance compared to Random across all models, except for ConvNet. The described methods collaboratively yield a condensed image that is impactful without inducing over-condensation, particularly evident in the case of ConvNet. This trend becomes more pronounced as IPC decreases. Simultaneously, our method competes favorably with other methods on ConvNet. Noteworthy is that HMDC performs the best on IPC 1 for all models. This suggests that the proposed HMDC effectively captures general features and incorporates them into a limited synthetic image. Unlike previous methods, HMDC shows promise for training large models from images compressed from relatively small models, aligning with the goal of dataset condensation. Despite utilizing two models, HMDC requires only 100 iterations and consumes less time than other models, typically using 1,200 to 20,000 iterations.


\label{sec:exp:ablation}
\subsection{Ablation Studies}

We conducted an ablation study to evaluate the impact of the proposed features on performance. Specifically, we examined performance by removing Mutual Distillation by Spatial-Sementic Decomposition and the Gradient Balance Module. Table \ref{table:ablation} presents the experimental result of ablation. The experimental results demonstrate that each element contributes to performance, and when both methods are employed, they exhibit synergy.

In the Ablation Study, the results demonstrate that the isolated application of the Gradient Balance Module (GBM) and Mutual Distillation (MD) can be beneficial in certain scenarios, though they tend to favor either Convolutional Neural Network (CNN) model. This leads to good performance on certain models and poor performance on others. The isolated use of the GBM tends to favor smaller models because the semantic distance between the two models fails to encapsulate the complex patterns necessary for larger models. Conversely, when only MD is employed, it extracts more generalized features, performing well with larger models. However, the results are biased due to the gradient difference in the condensed model. By combining these methods, a balanced gradient is achieved for each model, significantly reducing the semantic distance between them. This synergy not only enhances the overall performance but also ensures a more model-agnostic improvement.

\begin{table}[t!]
\centering
\caption{Table illustrating the outcomes of the ablation study. GBM means Gradient Balance Module and MD means Mutual Distillation by Spatial-Semantic Decomposition. The results demonstrate the individual contributions of the presented factors to performance enhancements, revealing a synergistic effect when employing them simultaneously.}
\scalebox{0.59}{
\begin{tabular}{ScSc|ScScScScSc:ScScScSc:Sc}
\hline
\multicolumn{1}{Sc}{\multirow{3}{*}{GBM}} & \multicolumn{1}{Sc|}{\multirow{3}{*}{MD}} & \multicolumn{10}{Sc}{CIFAR-10 (IPC 10)} \\ \cline{3-12} 
&&
\multirow{2}{*}[-.75pt]{\begin{tabular}[c]{@{}c@{}}ConvNet\\(0.3M)\end{tabular}} &
\multirow{2}{*}[-.75pt]{\begin{tabular}[c]{@{}c@{}}ResNet18\\(11M)\end{tabular}} &
\multirow{2}{*}[-.75pt]{\begin{tabular}[c]{@{}c@{}}ResNet50\\(22M)\end{tabular}} &
\multirow{2}{*}[-.75pt]{\begin{tabular}[c]{@{}c@{}}ResNet101\\(43M)\end{tabular}}&
\multirow{2}{*}[-.75pt]{\begin{tabular}[c]{@{}c@{}}CNN\\Average\end{tabular}} &
\multirow{2}{*}[-.75pt]{\begin{tabular}[c]{@{}c@{}}ViT-tiny\\(5.5M)\end{tabular}} &
\multirow{2}{*}[-.75pt]{\begin{tabular}[c]{@{}c@{}}ViT-small\\(21M)\end{tabular}} &
\multirow{2}{*}[-.75pt]{\begin{tabular}[c]{@{}c@{}}ViT-base\\(86M)\end{tabular}} &
\multirow{2}{*}[-.75pt]{\begin{tabular}[c]{@{}c@{}}ViT\\Average\end{tabular}} &
\multirow{2}{*}[-.75pt]{\begin{tabular}[c]{@{}c@{}}Average\end{tabular}} \\
&&&&&&&&& \\ \hline
\multicolumn{2}{Sc|}{DREAM}& \textbf{49.23±0.16} & 57.97±2.77 & 67.85±2.54 & 64.12±3.95 & 59.79±2.36 & 55.40±6.06 & 71.77±8.74 & 68.88±8.16 & 65.35±7.65 & 62.18±4.63 \\\hline
&& 46.42±0.37 & \underline{72.11±0.06} & 76.46±1.09 & 76.85±0.88 & 67.96±0.60 & \textbf{74.09±4.36} & \textbf{90.01±1.00} & \underline{82.71±0.76} & \underline{82.27±2.04} & \underline{74.09±1.22} \\
\checkmark&& 47.43±0.46 & 71.62±1.25 & 76.92±0.36 & 77.70±1.77 & 68.41±0.96 & 73.67±2.38 & 86.62±3.11 & 78.53±6.77 & 79.61±4.08 & 73.21±2.30 \\
&\checkmark& 47.30±0.41 & \textbf{72.01±1.46} & \textbf{78.05±2.23} & \underline{79.07±1.69} & \underline{69.11±1.45} & 70.96±4.02 & 86.25±5.25 & 80.16±5.36 & 79.12±4.88 & 73.40±2.92 \\\hline
\checkmark&\checkmark& \underline{47.54±82.73} & 69.75±75.10 & \underline{77.99±0.73} & \textbf{82.25±0.34} & \textbf{69.38±1.53} & \underline{73.60±0.93} & \underline{89.02±0.88} & \textbf{85.58±4.20} & \textbf{82.73±1.42} & \textbf{75.10±1.77} \\\hline
\end{tabular}
}
\label{table:ablation}
\end{table}

\begin{table}[t!]
\caption{
The performance variations across model combinations, depicting results for pairs of identical models (CNN + CNN) and pairs involving larger models than those employed in the experiment.
}
\resizebox{1.0 \textwidth}{!}{
\begin{tabular}{cccc|ccccc:cccc:c}
\hline
\multicolumn{1}{Sc}{\multirow{3}{*}{ConvNet}} & \multicolumn{1}{Sc}{\multirow{3}{*}{ResNet18}} & \multicolumn{1}{Sc}{\multirow{3}{*}{ViT-tiny}} & \multicolumn{1}{Sc}{\multirow{3}{*}{ViT-small}} & \multicolumn{10}{|Sc}{CIFAR-10 (IPC 10)} \\ \cline{5-14} 
&&&&
\multirow{2}{*}[-.75pt]{\begin{tabular}[c]{@{}c@{}}ConvNet\\(0.3M)\end{tabular}} &
\multirow{2}{*}[-.75pt]{\begin{tabular}[c]{@{}c@{}}ResNet18\\(11M)\end{tabular}} &
\multirow{2}{*}[-.75pt]{\begin{tabular}[c]{@{}c@{}}ResNet50\\(22M)\end{tabular}} &
\multirow{2}{*}[-.75pt]{\begin{tabular}[c]{@{}c@{}}ResNet101\\(43M)\end{tabular}}&
\multirow{2}{*}[-.75pt]{\begin{tabular}[c]{@{}c@{}}CNN\\Average\end{tabular}} &
\multirow{2}{*}[-.75pt]{\begin{tabular}[c]{@{}c@{}}ViT-tiny\\(5.5M)\end{tabular}} &
\multirow{2}{*}[-.75pt]{\begin{tabular}[c]{@{}c@{}}ViT-small\\(21M)\end{tabular}} &
\multirow{2}{*}[-.75pt]{\begin{tabular}[c]{@{}c@{}}ViT-base\\(86M)\end{tabular}} &
\multirow{2}{*}[-.75pt]{\begin{tabular}[c]{@{}c@{}}ViT\\Average\end{tabular}} &
\multirow{2}{*}[-.75pt]{\begin{tabular}[c]{@{}c@{}}Average\end{tabular}} \\
&&&&&&&&& \\ \hline
 \multicolumn{4}{Sc|}{Random}
 & 36.45±0.12 & 56.59±4.86 & {69.55±9.69} & {82.03±2.06} & {61.15±4.18} & {59.41±9.20} & \underline{90.11±1.25} & {81.26±5.66} & {76.93±5.37} & {67.91±4.69} \\\hline
\checkmark&&\checkmark& &\textbf{47.54±0.73} & \underline{69.75±0.34} & \underline{77.99±1.53} & \underline{82.25±0.93} & \textbf{69.38±0.88} & \underline{73.60±4.20} & 89.02±1.42 & \underline{85.58±1.77} & \underline{82.73±2.46} & \underline{75.10±1.56} \\
\checkmark&\checkmark&&\multicolumn{1}{c|}{} & \underline{45.85±0.71} & 44.09±4.33 & 47.32±1.06 & 32.91±2.31 & 42.54±2.10 & 33.79±1.90 & 34.93±5.55 & 34.58±14.63 & 34.43±7.36 & 39.07±4.35 \\
&\checkmark&&\checkmark& 38.10±0.40 & \textbf{70.67±0.93} & \textbf{79.85±1.59} & \textbf{83.82±1.26} & \underline{68.11±1.05} & \textbf{75.80±2.54} & \textbf{92.42±0.46} & \textbf{88.60±4.28} & \textbf{85.61±2.43} & \textbf{75.61±1.64} \\
\hline
\end{tabular}
}
\vspace{1mm}
\label{tab:models}
\end{table}

\begin{table}[t!]
\caption{Comparison table between simply using ViT-Tiny and using the presented method.}
\centering
\scalebox{0.52}{
\begin{tabular}{Sc|ScSc|ScScScScSc:ScScScSc:Sc}
\hline
\multicolumn{1}{Sc|}{\multirow{3}{*}{Methods}} & \multicolumn{1}{Sc}{\multirow{3}{*}{ConvNet}} & \multicolumn{1}{Sc|}{\multirow{3}{*}{ViT-tiny}} & \multicolumn{10}{Sc}{CIFAR-10 (IPC 10)} \\ \cline{4-13} 
&&&
\multirow{2}{*}[-.75pt]{\begin{tabular}[c]{@{}c@{}}ConvNet\\(0.3M)\end{tabular}} &
\multirow{2}{*}[-.75pt]{\begin{tabular}[c]{@{}c@{}}ResNet18\\(11M)\end{tabular}} &
\multirow{2}{*}[-.75pt]{\begin{tabular}[c]{@{}c@{}}ResNet50\\(22M)\end{tabular}} &
\multirow{2}{*}[-.75pt]{\begin{tabular}[c]{@{}c@{}}ResNet101\\(43M)\end{tabular}}&
\multirow{2}{*}[-.75pt]{\begin{tabular}[c]{@{}c@{}}CNN\\Average\end{tabular}} &
\multirow{2}{*}[-.75pt]{\begin{tabular}[c]{@{}c@{}}ViT-tiny\\(5.5M)\end{tabular}} &
\multirow{2}{*}[-.75pt]{\begin{tabular}[c]{@{}c@{}}ViT-small\\(21M)\end{tabular}} &
\multirow{2}{*}[-.75pt]{\begin{tabular}[c]{@{}c@{}}ViT-base\\(86M)\end{tabular}} &
\multirow{2}{*}[-.75pt]{\begin{tabular}[c]{@{}c@{}}ViT\\Average\end{tabular}} &
\multirow{2}{*}[-.75pt]{\begin{tabular}[c]{@{}c@{}}Average\end{tabular}} \\
&&&&&&&&& \\ \hline
Random &&& 36.45±0.12 & 56.59±4.86 & \underline{69.55±9.69} & \textbf{82.66±1.82} & \underline{61.31±4.12} & 59.36±9.19 & \textbf{90.11±1.25} & \underline{81.26±5.66} & \underline{76.91±5.37} & \underline{68.00±4.66} \\
Dream &\checkmark&& \textbf{49.23±0.16} & \underline{57.97±2.77} & 67.85±2.54 & 64.12±3.95 & 59.79±2.36 & 55.40±6.06 & 71.77±8.74 & 68.88±8.16 & 65.35±7.65 & 62.18±4.63 \\
Dream &&\checkmark& 25.77±0.57 & 49.28±2.70 & 56.65±4.37 & 62.26±4.08 & 48.49±2.93 & \underline{60.26±3.78} & 62.68±10.5 & 51.77±22.3 & 58.24±12.2 & 52.67±6.91\\\hline
HMDC &\checkmark&\checkmark& \underline{47.54±0.73} & \textbf{69.75±0.34} & \textbf{77.99±1.53} & \underline{82.25±0.93} & \textbf{69.38±0.88} & \textbf{73.60±4.20} & \underline{89.02±1.42} & \textbf{85.58±1.77} & \textbf{82.73±2.46} & \textbf{75.10±1.56} \\\hline
\end{tabular}
}
\vspace{1mm}
\label{tab:dependancy}
\end{table}

\begin{figure}[t] 
    \centering
    \vspace{3mm}
    \caption{Comparision of condensed images between DREAM \cite{Liu_2023_ICCV} and HMDC(Ours)}
    \begin{subfigure}[h]{0.48\textwidth}
    \includegraphics[width=\textwidth]{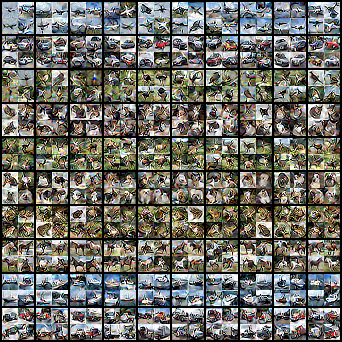}
    \caption{Condensed image generated by DREAM \cite{Liu_2023_ICCV} method, with CIFAR10, 10 images per class.}
    \vspace{3mm}
    \label{fig:dream}
    \end{subfigure}
    \centering
    \begin{subfigure}[h]{0.48\textwidth}
    \includegraphics[width=\textwidth]{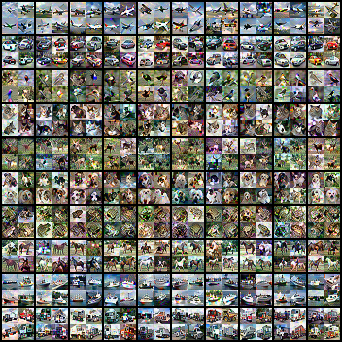}
    \caption{Condensed image generated by Heterogeneous Model Dataset Condensation (HMDC) method, with CIFAR10, 10 images per class.}
    \label{fig:ours}
    \end{subfigure}
\end{figure}

\subsection{Qualitative Results}

Figure \ref{fig:dream} presents a condensed image generated using the DREAM \cite{Liu_2023_ICCV} method.
In contrast, Figure \ref{fig:ours} depicts a condensed image created using the Heterogeneous Model Dataset Condensation approach.
Overall, we observe that objects exhibit the same characteristics, such as increased contrast and sharpened edges.
While the visual disparities are minimal, it is noteworthy that the image generated by our method exhibits fewer artifacts and distortions compared to DREAM.
These distortions appear to be caused by over-condensation, improving performance on certain models but degrading performance on others.

\subsection{Analysis}
Table \ref{tab:models} illustrates the performance outcomes of Heterogeneous Model Dataset Condensation across various model combinations.
Significantly, there is an evident decrease in overall performance when the CNN is used uniformly, contrary to the intended objective of the method.
In the third row, we adjusted the learning rate of the affine layer and layer-matching matrix to $0.001$. This suggests that with minimal modification, larger models can be accommodated using HMDC. We can also see that the use of larger models shows an overall benefit.

Table \ref{tab:dependancy} presents a comparative analysis of our experimental results with the state-of-the-art method, DREAM, changing the model into ViT-Tiny.
The results reveal an overall enhancement in performance for ViT-Tiny attributed to an increased understanding of the model.
However, overall performance drops, especially in the ConvNet and it is notably worse compared to using random images. This trend is further emphasized by the large gap between HMDC in combination with ConvNet.
Conversely, in ConvNet, overall performance is diminished due to the model's limited capacity, with a pronounced decline in the ViT series. This observation underscores the model dependency of the condensed image and the condensed image is over-condensed.

\begin{figure}[t]
    \centering
    \caption{A logarithmic plot depicting the gradient magnitude evolution of a synthetic image throughout the training process. L1 and L2 refer to the optimization targets in Eq. \ref{eq:dual}. L3 is $\mathrm{MSE}\left(\nabla\mathcal{L}_{\mathrm{MD}}(\textbf{x}^t),\nabla\mathcal{L}_{\mathrm{MD}}\right)$. }
     \includegraphics[width=0.8\textwidth]{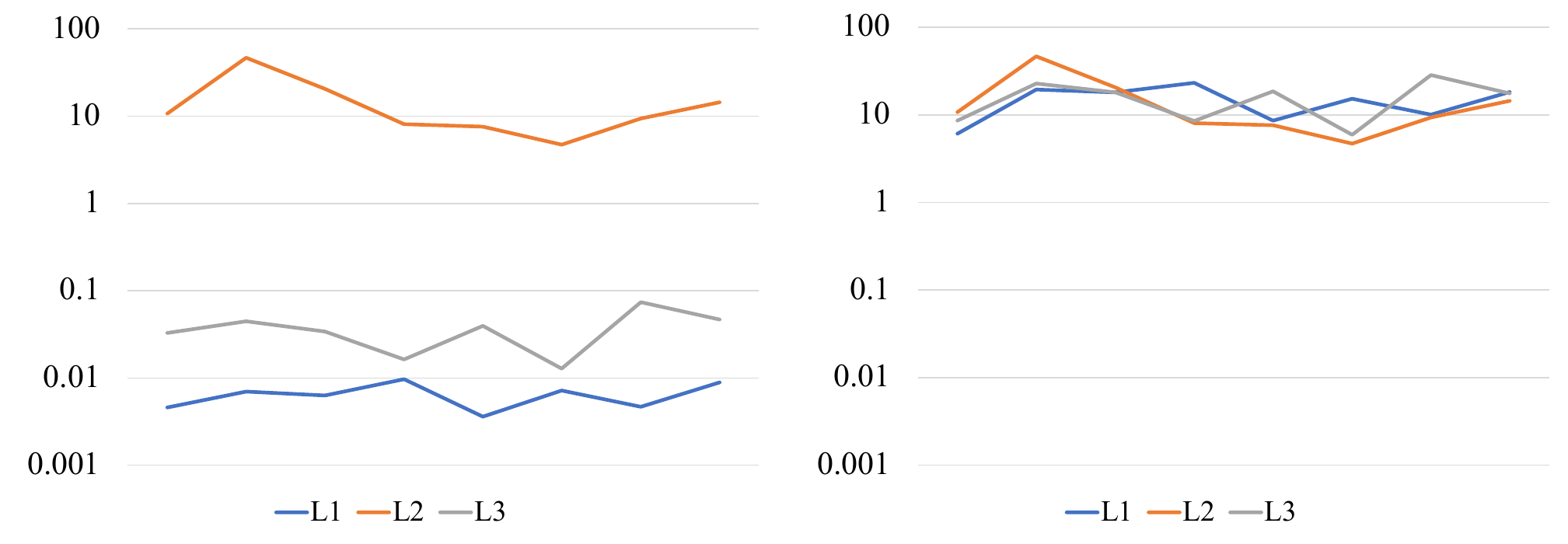}
    \label{fig:grad}
\end{figure}

Figure \ref{fig:grad} depicts the maximum gradient magnitude of the synthetic image after the learning step.
On the left, before integrating the Gradient Balance Module, there is a substantial disparity in the gradient magnitudes, leading to the neglect of other losses.
Following the inclusion of the Gradient Balance Module, the gradient magnitudes from each loss function become uniform.
This ensures balanced dataset condensation irrespective of the model, underscoring the essential role of the Gradient Balance Module in Heterogeneous Model Dataset Condensation.

\section{Conclusion}
\label{sec:conclusion}
We identified a model dependency issue in existing dataset condensation methods and proposed a remedy by balancing different heterogeneous models. However, employing two distinct models for dataset condensation introduces two challenges: bias arising from differences in gradient magnitude between the models and semantic distance resulting from the models converging to their respective optimal points. To address the gradient magnitude disparity, we introduced the Gradient Balance Module. To tackle the semantic distance issue, we proposed Mutual Distillation by Spatial-Semantic Decomposition. The combination of these components effectively alleviated the model dependency problem.

However, there are still some limitations. Firstly, it is impossible to surpass the model's capacity limits, as evidenced by the performance gap for ViT-Base due to a fifteenfold difference in parameter number. Furthermore, due to the continued use of floating points in condensed images, they occupy four times the capacity of the equivalent number of real images. From this standpoint, there is a need for quantization-aware condensation that enables the synthetic image to be treated like a real image.
\section* {Acknowledgement}

This work was supported by MSIT (Ministry of Science and ICT), Korea, under the ITRC (Information Technology Research Center) support program (IITP-2024-RS-2023-00258649) supervised by the IITP (Institute for Information \& Communications Technology Planning \& Evaluation), and in part by the IITP grant funded by the Korea Government (MSIT) (Artificial Intelligence Innovation Hub) under Grant 2021-0-02068, and by the IITP grant funded by the Korea government (MSIT) (No.RS-2022-00155911, Artificial Intelligence Convergence Innovation Human Resources Development (Kyung Hee University)).

%
%
\bibliographystyle{splncs04}
\bibliography{main}
\end{document}


\title{Supplimentary Material for Towards Model-Agnostic Dataset Condensation by Heterogeneous Models} 

\titlerunning{Heterogeneous Models Dataset Condensation}

\authorrunning{J.-Y. Moon et. al.}


\appendix
\vspace{-15mm}
\begin{figure}[h!]
    \centering
    \begin{subfigure}[t]{0.48\textwidth}
        \includegraphics[width=\textwidth]{figure/CIFAR10IPC1.pdf}
        \caption{The results obtained from CIFAR-10 \cite{krizhevsky2009learning} at IPC 1.}
        \label{fig:CIFAR10IPC1}
    \end{subfigure}
    \centering
    \begin{subfigure}[t]{0.48\textwidth}
        \includegraphics[width=\textwidth]{figure/CIFAR10IPC10.pdf}
        \caption{The results obtained from CIFAR-10 at IPC 10.}
        \label{fig:CIFAR10IPC10}
    \end{subfigure}
\caption{The graph depicts the performance gap between Random Data and condensation using each method. In most instances, images generated through these methods show a decrease in performance, except for ConvNet. Conversely, Heterogeneous Model Dataset Condensation (HMDC) demonstrates consistent performance across various models.}
\label{fig:CIFAR10}
\end{figure}

\vspace{-15mm}

\begin{figure}[h!]
    \centering
    \begin{subfigure}[t]{0.48\textwidth}
        \includegraphics[width=\textwidth]{figure/CIFAR100IPC1.pdf}
        \caption{The results obtained from CIFAR-100 at IPC 1.}
        \label{fig:CIFAR10IPC1}
    \end{subfigure}
    \centering
    \begin{subfigure}[t]{0.48\textwidth}
        \includegraphics[width=\textwidth]{figure/CIFAR100IPC10.pdf}
        \caption{The results obtained from CIFAR-100 at IPC 10.}
        \label{fig:CIFAR10IPC10}
    \end{subfigure}
\caption{The graph depicts the performance gap between Random Data and condensation using each method. In most instances, images generated through these methods show a decrease in performance, except for ConvNet. Conversely, Heterogeneous Model Dataset Condensation (HMDC) demonstrates consistent performance across various models.}
\label{fig:CIFAR100}
\end{figure}
\vspace{-15mm}

\section{Visualization of Comparison with Random}

We have translated the tabulated data presented in Table \textcolor{red}{1} of the main text into a visual representation depicted in Figure \textcolor{red}{1}. This graphical representation vividly illustrates the escalating impact of model dependency, a phenomenon particularly noticeable as the number of Images Per Class (IPC) decreases. Clearly, a majority of the generated images tend to exhibit bias towards Convolutional Neural Network (CNN) \cite{lecun1998gradient} and showcase suboptimal performance compared to random selection, posing a significant hindrance to their usability. However, it is noteworthy that Heterogeneous Model Dataset Condensation (HMDC) significantly alleviates this issue, thereby expanding the applicability and robustness of Dataset Condensation (DC). This visual insight contributes to a more nuanced understanding of the intricate dynamics of model dependency in the context of dataset condensation.

\input{table/Additional_Main}
\section{Additional Experiments on Other Datasets}

We conducted experiments on CIFAR-100 \cite{krizhevsky2009learning} and TinyImageNet \cite{le2015tiny} to validate the effectiveness of Heterogeneous Model Dataset Condensation on datasets other than CIFAR-10.

Figure \ref{fig:CIFAR100} illustrates the performance contrast with Random Image on CIFAR-100, in relation to Figure \ref{fig:CIFAR10}.
As a comparison group, we used CAFE, IDM, IDC, and DREAM as in the main text.
Once more, it is evident that HMDC exhibits commendable performance across the board, and distinctively, it avoids the model dependency challenges experienced by other methods.
The results of these experiments are presented in Table \ref{tab:additional_exp}.

We observe that CIFAR-100 exhibits strong performance across various models, excluding ConvNet, mirroring the patterns seen in CIFAR-10. Images generated at IPC 1 displayed notably low perceptual quality, yet remained valuable for training purposes. Although this may have influenced generalization performance to some extent, the overall performance distribution appears to be independent of the specific model employed.

We conducted an experiment to assess the performance of the proposed method on TinyImageNet.
However, owing to the heightened computational requirements resulting from the increased resolution and number of classes, we opted for a partial training approach to discern trends.
Despite the limited training, Dream was trained more in proportion and showing performance improvement, effectively illustrating the observed trends. Notably, Dream continued to exhibit model dependency on the larger dataset. In contrast, HMDC demonstrated the ability to mitigate performance disparities between models, showcasing a more consistent improvement across different model architectures.

\begin{table}[t]
\centering
\begin{minipage}[h]{.48\textwidth}
\resizebox{\columnwidth}{!}{
    \begin{tabular}{ccc}
    \hline
    Methods  & SimpleCNN   & HuBERT \\ \hline
    Dream    & 10.00±0.00$^*$ & 12.08±1.19 \\
    \textbf{HMDC (Ours)} & \textbf{17.08±1.91} & \textbf{12.50±2.50} \\
    \hline
    \end{tabular}
}	
\captionof{table}{Table of results from experimenting with HMDC and Dream in audio modality. $*$ means failure to converge.}
\label{tab:audio}
\end{minipage}
\centering
\begin{minipage}[h]{.48\textwidth}
\resizebox{\columnwidth}{!}{
    \begin{tabular}{ccc}
    \hline
    Methods  & SimpleRNN & BERT \\ \hline
    Dream    & 50.00±0.00$^*$ & 50.97±0.82 \\
    \textbf{HMDC (Ours)} & 50.00±0.00$^*$ & \textbf{51.12±0.02}  \\
    \hline
    \end{tabular}
}
\captionof{table}{Table of results from experimenting with HMDC and Dream in language modality. $*$ means failure to converge.}
\label{tab:language}
\end{minipage}
\end{table}

\section{Additional Modality Experiments}
We extended our HMDC initially designed for the image domain to other audio and language domains.
For each domain, ESC-10 \cite{piczak2015dataset} (10 classes) in audio and the IMDB Dataset \cite{maas-etal-2011-learning} (2 classes) in text served as compression targets into $1$ sample per class.
We defined SimpleCNN (6 layers) and SimpleLSTM (6 layers) as correspondence for ConvNet, which is predominantly used in DC, respectively.
Additionally, HuBERT \cite{hsu2021hubert} and BERT, structurally similar to ViT \cite{dosovitskiy2020image}, were employed as secondary models to maintain heterogeneity. 
In each scenario, we applied the method of interpolating the temporal feature corresponding to the spatial feature.
Tables \ref{tab:audio} and \ref{tab:language} present the comparative results, demonstrating that our HMDC has the potential for application across various modalities.

\section{Additional Implementation Details}

We ran all of our experiments on the A5000 GPU. In Table \textcolor{red}{1} and \textcolor{red}{2} on the main paper, we re-implemented CAFE \cite{wang2022cafe}, IDM \cite{zhao2023improved}, IDC \cite{kim2022dataset}, and DREAM \cite{Liu_2023_ICCV} based on their official implementations to enable model and seed flexibility. Each implementation prioritizes adherence to the paper's representation, with the official code considered secondarily. The computationally intensive experiments on TinyImageNet lasted a quarter of the iterations of the other experiments. We checked that each method exhibits comparable performance to the official code. For the evaluation process, we followed IDC's augmentation strategy, incorporating color adjustments, cropping, and CutMix \cite{yun2019cutmix}. Subsequently, we measured the performance on the test data after training for $1,000$ epochs at the specified learning rate for each model mentioned in the text, without any learning rate scheduling.

In the experiments detailed in Table \textcolor{red}{4} of the main text, when we applied Vision Transformer (ViT) \cite{dosovitskiy2020image} to DREAM, the image failed to converge, necessitating a reduction in the learning rate of the image from $5e^{-2}$ to $5e^{-5}$, as outlined in the original paper.

In Figure \textcolor{red}{5} on the main paper, we adhered to HMDC's training protocol and examined the Maximum Gradient Value both with and without the Gradient Balance Module at each of the 10 steps, conducting a total of 8 iterations.

We applied a learning rate of $1e^{-3}$ for the affine layer and layer-matching matrix when utilizing ViT-Small. Conversely, for all other cases, a learning rate of $1e^{-1}$ was employed. Additionally, to accommodate memory constraints, the batch sizes were set as $128$, $16$, $32$, $16$, $4$, $128$, $128$, $64$, and $4$, respectively, starting after Random.
%
%
\bibliographystyle{splncs04}
\bibliography{main}